\documentclass[letterpaper, 10 pt, conference]{ieeeconf}

\IEEEoverridecommandlockouts
\usepackage{graphicx}
\usepackage{amsmath}
\usepackage{amssymb}
\usepackage{booktabs}
\usepackage{xcolor}
\usepackage{fancyhdr}
\fancypagestyle{preprint}{%
  \fancyhf{}                          
  \fancyfoot[C]{\footnotesize Preprint. Under review.}
}
\definecolor{linkblue}{RGB}{0,51,153}
\usepackage[colorlinks=true, allcolors=linkblue,
  pdftitle={Mask2Real-WM: Segmentation Masks as a Sim-to-Real Bridge for Controllable Dexterous World Models},
  pdfauthor={}]{hyperref}

\makeatletter
\long\def\@makecaption#1#2{%
\ifx\@captype\@IEEEtablestring%
\begin{center}{\footnotesize #1}\\[-2pt]{\footnotesize\scshape #2}\end{center}%
\@IEEEtablecaptionsepspace%
\else
\@IEEEfigurecaptionsepspace%
\setbox\@tempboxa\hbox{\footnotesize #1.~~ #2}%
\ifdim \wd\@tempboxa >\hsize%
\setbox\@tempboxa\hbox{\footnotesize #1.~~ }%
\parbox[t]{\hsize}{\footnotesize \noindent\unhbox\@tempboxa#2}%
\else%
\ifcenterfigcaptions \hbox to\hsize{\footnotesize\hfil\box\@tempboxa\hfil}%
\else \hbox to\hsize{\footnotesize\box\@tempboxa\hfil}%
\fi\fi\fi}
\makeatother

\newcommand{\modelname}{Mask2Real-WM}
\newcommand{\msr}{Cascade-SR}
\newcommand{\mro}{Cascade-R}
\newcommand{\mso}{Cascade-S}
\newcommand{\monosr}{Mono-SR}
\newcommand{\monoro}{Mono-R}
\newcommand{\wmone}{WM1}
\newcommand{\wmtwo}{WM2}

\title{\LARGE \bf
Mask2Real-WM: Controllable Dexterous World Models via Segmentation Masks as a Sim-to-Real Bridge
}

\author{Riccardo O.~Feingold$^{1}$, Davide Liconti$^{1}$, Chenyu Yang$^{1}$, and
Robert K.~Katzschmann$^{1}$
\thanks{$^{1}$Soft Robotic Lab, Department of Mechanical and Process Engineering,
        ETH Zurich, Switzerland
        {\tt\small rfeingold@ethz.ch}}%
}
\begin{document}

\maketitle
\thispagestyle{preprint}   
\pagestyle{preprint}       

\begin{abstract}
Learning action-conditioned world models for dexterous manipulation that are genuinely controllable requires capturing complex, high-dimensional hand kinematics from limited real-world data. We present \modelname{}, a two-stage world model that improves controllability for dexterous hands under a limited real-data budget by decoupling pixel prediction into a dynamics model and a rendering model. The dynamics model predicts future segmentation masks from past masks and a high-dimensional action sequence. The rendering model converts the predicted masks into photorealistic RGB images. This design allows us to train the dynamics model on a large synthetic dataset spanning the full range of hand motions and interactions. We then fine-tune the dynamics model and train the rendering model on only 2.5 h of real demonstrations to obtain a controllable world model. We compare five models with matched training recipes on a 23-DoF robotic system,  measuring per-DoF controllability both with random target commands and with sinusoidal per-DoF actuation,  judged blind by human raters.
The decoupled model achieves the strongest controllability, with further gains from simulation data.
Beyond improved controllability, \modelname{} produces sharper frames while maintaining comparable perceptual quality.
\end{abstract}


\section{Introduction}
\label{sec:introduction}

World models are increasingly used in robot learning~\cite{hou2026worldmodelsurvey}. In this setting, they commonly take two forms: policy backbones (World Action Models) that jointly predict actions and observations, and \emph{action-conditioned} world models that predict future states given actions. The latter, provided the model truly follows the actions (i.e., is \emph{controllable}), support policy evaluation~\cite{guo2026ctrlworld}, planning~\cite{goswami2026dexwm}, and dreaming~\cite{hafner2025dreamerv3}.

\begin{figure}[!t]
  \centering
  \includegraphics[width=0.95\linewidth]{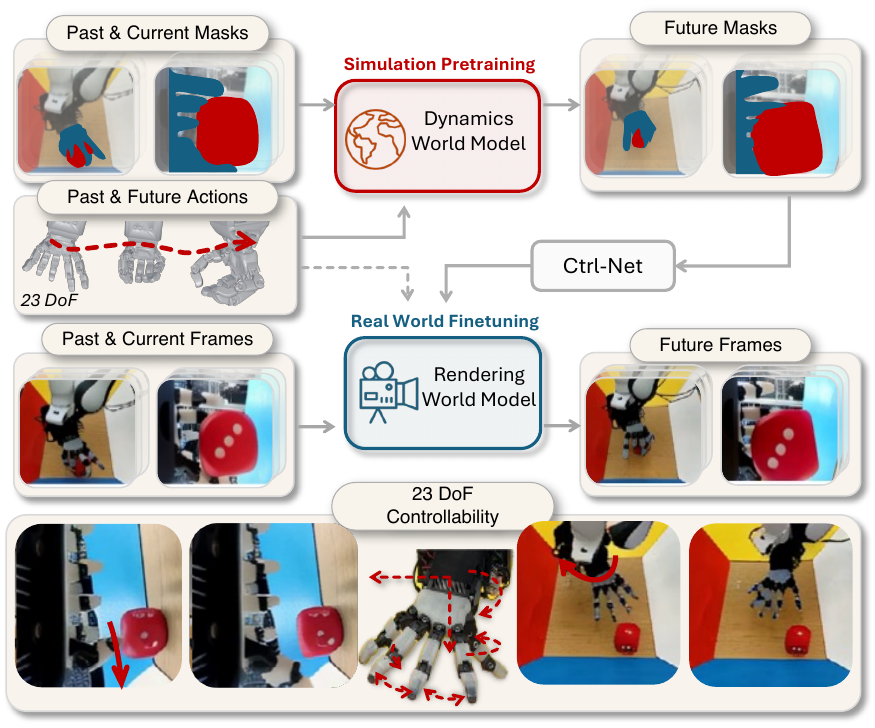}
  \caption{\textbf{\modelname{}.} A dynamics world model predicts future segmentation masks from past masks and the 23-D action sequence and is midtrained on synthetic simulation data; a rendering world model paints photorealistic RGB onto the predicted masks and is trained on 2.5\,h of real world interactions. \modelname{} faithfully predicts each DoF’s commanded motion, including motion patterns poorly represented in real demonstrations.}
  \label{fig:teaser_top}
\end{figure}

However, training controllable action-conditioned world models for dexterous hands~\cite{christoph2025orca} is challenging for several reasons. First, conditioning on the high-dimensional action space (23-D in our setup: a 6-D Cartesian end-effector pose and 17-D hand-joint positions) is more demanding than for parallel-jaw grippers: the model must capture the response to each action dimension. Limited task demonstrations may predominantly show joints moving together, encouraging learned responses that couple their motions even when only one joint is commanded to move. Second, generated videos must resolve fine finger motion and interactions clearly. Limited availability and slow collection of dexterous-hand data further constrain the training of these models.

We present a method that addresses the controllability challenge for dexterous hands. We introduce \modelname{} (Fig.~\ref{fig:teaser_top}), a two-stage action-conditioned world model that decouples pixel prediction into an action-conditioned \emph{dynamics model} predicting future segmentation masks, and a separate \emph{rendering model} that paints RGB onto those masks. The smaller sim-to-real gap in mask space enables the dynamics model to benefit from large-scale midtraining on 53.7\,h of synthetic simulation data, followed by fine-tuning on just 2.5\,h of real demonstrations. This decomposition reframes the problem as two more tractable subproblems, the first of which can effectively use synthetic simulation data.

This yields higher controllability over the hand action space: the predicted videos follow the commanded actions more closely. Our platform is a robotic arm and dexterous hand, with data collected from a continuous pick-and-place task and predictions made for two camera views, workspace and wrist. We evaluate controllability with two experiments: first, blind A/B comparisons against monolithic baselines on random whole-pose targets; second, rater scoring on individually actuated fingers, assessing whether the commanded degree of freedom moves independently without coupling to its neighbors.
Raters strongly prefer \modelname{} over monolithic models trained with matching recipes, in both experiments. Simulation midtraining of the mask dynamics model further improves controllability in both experiments. For the monolithic model, the same simulation data instead hurts controllability, likely due to distribution shift between simulated and real RGB.

These gains come without compromising perceptual quality, which remains on par with or better than the baselines, while \modelname{} consistently produces sharper predictions.
\begin{figure*}[!t]
  \vspace*{4pt}
  \centering
  \includegraphics[width=0.85\textwidth]{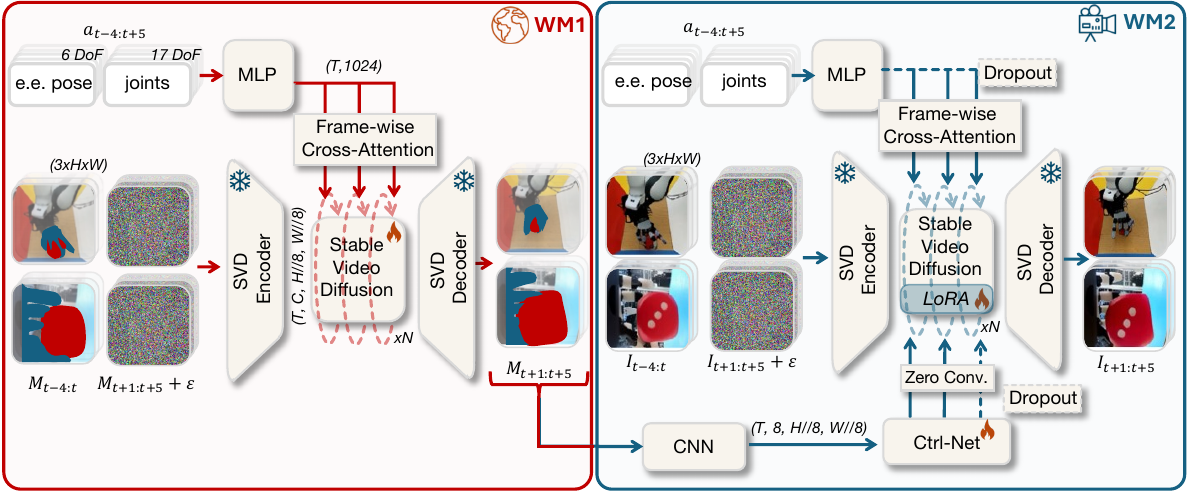}
  \caption{\textbf{\modelname{} architecture.} WM1: An action-conditioned dynamics model denoises future segmentation masks from past masks and the action sequence; WM2: predicted future masks are fed via a ControlNet into a LoRA-finetuned SVD model, which diffuses future frames conditioned on past RGB and both past and future actions; masks and actions are subject to dropout during training for regularization.}
  \label{fig:method}
\end{figure*}

Our contributions are as follows:
\begin{enumerate}
\item We introduce \modelname{}, a two-stage action-conditioned world model that decouples pixel prediction into mask dynamics and RGB rendering and uses the mask space as a sim-to-real bridge.
\item We show substantial gains in whole-pose target following and independent per-DoF control on a 23-DoF robotic system: \modelname{} wins 75\% of non-tied human comparisons against the strongest monolithic model on random whole-pose targets, and scores higher on single-DoF command sweeps, with sharper frame predictions and without compromising perceptual quality on the evaluated videos.
\item We ablate training recipes and simulation data, showing that simulation training improves controllability in the decoupled model but reduces it in the monolithic model.
\end{enumerate}

\section{Related Work}
\label{sec:related_work}

\subsection{World Models for Robotic Manipulation}
Video-based world models learn visual and motion priors from large video corpora to predict how scenes evolve~\cite{bruce2024genie,agarwal2025cosmos}. In robotics, these predictions can support action selection~\cite{goswami2026dexwm,bi2026motus2} or serve as action-conditioned simulators for policy evaluation and planning~\cite{guo2026ctrlworld,kim2026dwm,bi2026motus2}. These models have also been applied to dexterous manipulation through RGB video prediction~\cite{kim2026dwm}, visual-feature prediction for planning~\cite{goswami2026dexwm}, and value estimation~\cite{bi2026motus2}. We focus on action-conditioned RGB simulation and evaluate how closely generated rollouts follow individual hand-joint commands.

\subsection{Structured Guidance for RGB Generation}
To improve the spatial accuracy and controllability of RGB generation, diffusion models incorporate structural guidance through auxiliary conditioning branches such as ControlNet~\cite{zhang2023controlnet}, visual action prompts~\cite{wang2025visualactionprompts}, and predicted~\cite{lou2026maskworldmodelpredicting} or rendered~\cite{chen2026bridgev2w} embodiment masks. These signals express geometry and motion in image coordinates, providing spatial guidance for the generated appearance. Mask World Model~\cite{lou2026maskworldmodelpredicting} is the closest concurrent work: it also learns to predict masks before generating RGB, but as a world-action model for a parallel-jaw gripper and without simulation midtraining. For manipulation, hand-mesh renderings ~\cite{kim2026dwm} and partially revealed robot videos~\cite{alzayer2026maskedvisualactions} have been proposed to condition the generation of interaction outcomes. Compared with approaches that supply rendered robot geometry as guidance, our model predicts both hand and object mask evolution before RGB synthesis. This mask evolution represents action-dependent changes in hand configuration and object state, making the guidance itself a learned dynamics prediction that can benefit from simulation midtraining.

\subsection{Action Conditioning and Controllability}
Spatial action interfaces seek to improve action following by expressing commands as embodiment masks~\cite{chen2026bridgev2w}, partially revealed robot videos~\cite{alzayer2026maskedvisualactions}, or image-space motion~\cite{li2026hydra0}. DexAC-WM~\cite{yuan2026dexacwm} addresses imbalanced high-DoF signals by preserving individual action dimensions through tokenization and aligning them with visual dynamics through local refinement and global modulation. Kim et al.~\cite{kim2026robotfactored} roll commands through a controller and kinematics to obtain nominal trajectories, then render robot geometry to condition scene-response prediction. However, their rendered guidance encodes nominal robot motion, leaving interaction dynamics to the RGB video model. We instead predict both the hand and object mask dynamics, using simulated data to improve action following and dynamics prediction independently of RGB appearance.

\section{Method}
\label{sec:models}

\subsection{Overview}
\label{sec:models:problem}
An action-conditioned world model predicts future observations from an observation history and a sequence of actions. For dexterous manipulation, these predictions should faithfully reflect the commanded hand motion while maintaining realistic visual appearance. Our approach (Fig.~\ref{fig:method}) separates these two requirements into mask dynamics prediction and RGB rendering. The dynamics stage predicts future hand and object segmentation masks from past masks and actions, and the rendering stage uses these predictions to guide RGB generation. This decomposition provides explicit spatial guidance for action following and allows the dynamics stage to learn from simulator-generated masks without requiring photorealistic simulation.

At discrete time step $t$, $I_t$ denotes the RGB observation, $m_t$ its segmentation mask, and $a_t$ the action command specifying the end-effector pose and hand-joint positions. The mask distinguishes the hand, object, and background. We denote the context length by $k$ and the prediction horizon by $H$.

\subsection{Mask Prediction}
\label{sec:models:mask2real}
For each prediction step in Fig.~\ref{fig:method}, the dynamics model (\wmone{}) first generates a future mask sequence, which then conditions the rendering model (\wmtwo{}). Let $p_{\mathrm{dyn}}$ and $p_{\mathrm{render}}$ denote their respective conditional distributions, and let $\mathbf{a}=a_{t-k+1:t+H}$ collect the actions over the context and prediction horizon. The two stages sample sequentially:
\begin{equation}
  \begin{aligned}
  m_{t+1:t+H} &\sim p_{\mathrm{dyn}}(\cdot \mid m_{t-k+1:t}, \mathbf{a}), \\
  I_{t+1:t+H} &\sim p_{\mathrm{render}}(\cdot \mid I_{t-k+1:t}, m_{t-k+1:t+H}, \mathbf{a}).
  \end{aligned}
\end{equation}
\wmone{} conditions only on the mask history and actions. \wmtwo{} receives the past RGB frames, the combined past and predicted mask sequence $m_{t-k+1:t+H}$, and the same actions. The predicted masks describe both hand and object evolution before RGB synthesis.

\subsection{Network Architecture and Training}
\label{sec:models:architecture}
Both stages use the Ctrl-World architecture~\cite{guo2026ctrlworld}, built on Stable Video Diffusion (SVD)~\cite{blattmann2023svd} and a pretrained variational autoencoder (VAE)~\cite{rombach2022ldm}. \wmone{} predicts mask sequences in the VAE latent space, conditioning on past mask latents and actions. \wmtwo{} uses past RGB latents as visual context and receives mask guidance through a convolutional encoder and a ControlNet branch~\cite{zhang2023controlnet}, extended to per-frame video conditioning. Both stages encode end-effector pose and hand-joint commands through separate action-encoder branches and inject their combined embeddings through frame-wise cross-attention.

Simulation supplies ground-truth masks for training \wmone{} on a broader range of hand motions than the real demonstrations cover (Section~\ref{sec:models:data}). We refer to this large-scale simulation training, before real-world fine-tuning, as \emph{midtraining}. In the full method, \wmone{} undergoes simulation midtraining followed by real-data fine-tuning, while \wmtwo{} is trained only on real data. This training strategy assigns simulation supervision to mask dynamics and real RGB supervision to appearance learning.

\subsection{Implementation Details}
\label{sec:models:implementation}
\label{sec:models:inference}
We use two camera views at $240{\times}135$ resolution, stacked along the image height. Masks are encoded as RGB images with three class colors: hand (green), object (red), and background (black). Synthetic masks are blurred with a Gaussian standard deviation of $\sigma{=}1.5$\,px to soften rasterization edges. Each action $a_t\in\mathbb{R}^{23}$ concatenates a 6-D end-effector pose and 17 hand-joint positions. The context length and prediction horizon are $k{=}H{=}5$ frames. The full method uses 53.7\,h of simulation trajectories for midtraining and 2.5\,h of real demonstrations for adaptation and rendering-model training.

We initialize the video backbones from SVD weights rather than from Ctrl-World's released checkpoint. The dual-branch action encoder replaces Ctrl-World's single MLP for a 7-D relative end-effector pose and gripper width. Pose and joint commands are normalized independently and projected by separate MLPs into 1024-dimensional per-frame embeddings, which are summed and injected into every U-Net block. \wmone{} concatenates past mask latents with noisy future mask latents along the time axis. For \wmtwo{}, the mask encoder is a three-block CNN (Conv2d--GroupNorm--SiLU, strided downsampling, transposed convolution, $1{\times}1$ convolution). Its ControlNet branch injects features into the backbone's encoder and middle blocks through zero-initialized convolutions. The backbone is adapted with LoRA~\cite{hu2022lora} (rank 16, $\alpha{=}16$); the CNN encoder and ControlNet are fully trained. The CNN consumes \wmone{}'s decoded mask images directly, without discretization. Mask and action conditional dropout (10\% each, 5\% both) enables classifier-free guidance. Training schedules and optimization settings are given in Section~\ref{sec:models:training}.

At inference, SAM~3~\cite{carion2026sam3} segments the initial five-frame real RGB context to seed the autoregressive rollout. Subsequent steps reuse the predicted masks and RGB frames as context, requiring no further segmentation.

\section{Experiments}
\label{sec:experiments}
\label{sec:evaluation}
\label{sec:results}

We evaluate \modelname{} against monolithic baselines~\cite{guo2026ctrlworld} along two axes: geometric controllability of the rendered hand, and video/perceptual quality. Controllability is assessed with two human rating experiments, an A/B test on randomly sampled target commands and a scoring experiment on individually actuated fingers via sine-wave commands, and with a mask-overlap metric (Section~\ref{sec:evaluation:geometry}). Perceptual quality is measured with standard metrics, complemented by a sharpness measure (Section~\ref{sec:evaluation:video}).

\subsection{Robot Platform and Datasets}
\label{sec:evaluation:setup}
\begin{figure}[ht]
  \centering
  \includegraphics[width=0.8\linewidth]{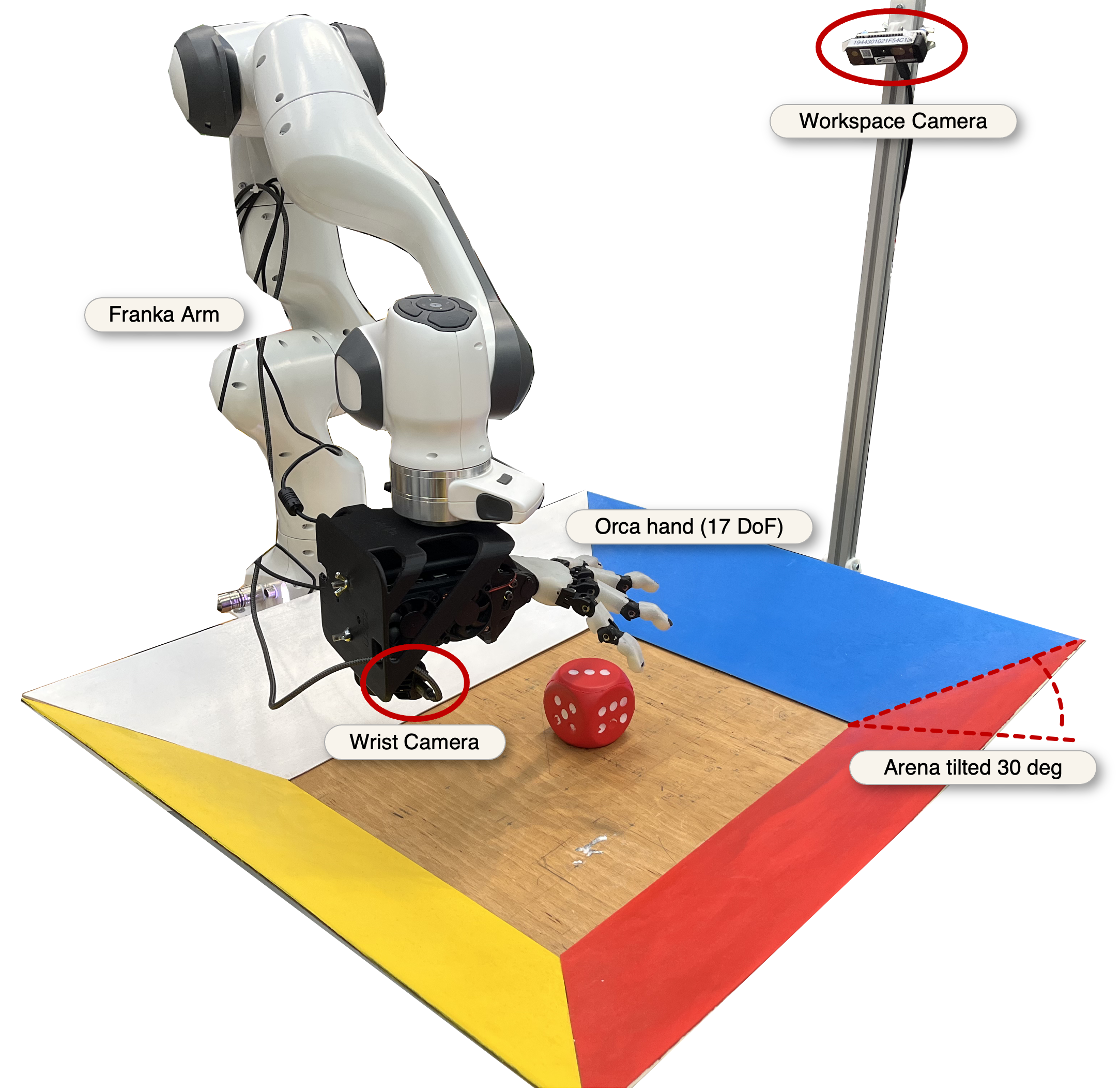}
  \caption{\textbf{Real-world rig.} Franka Panda arm with the ORCA hand in the arena, observed by a third-person and a wrist-mounted RGB camera.}
  \label{fig:robot_setup}
\end{figure}

The robot platform (Fig.~\ref{fig:robot_setup}) combines a 7-DoF Franka Emika Panda arm with the 17-DoF ORCA hand~\cite{christoph2025orca}. The model is commanded through a 23-dimensional action space (6-D Cartesian end-effector pose and 17 hand-joint positions). Data are collected in a custom arena with tilted, colored walls during pick-and-place and free-play interactions. Two RGB cameras provide observations: a static workspace view and a wrist-mounted camera.

The simulation counterpart uses the default Franka URDF and ORCA USD assets on the default Isaac Lab table, with arena geometry matched to physical measurements and no material, lighting, or texture tuning. Ground-truth masks come from the simulator's instance segmentation. To reduce the kinematic gap, we perform system identification following~\cite{bjelonic2025bridginggap}.

\label{sec:models:data}


The simulation corpus comprises 13,950 Isaac Lab~\cite{mittal2023isaaclab} episodes with ground-truth masks (53.7\,h at 5\,fps) over five objects, including a cube like the real one. MimicGen~\cite{mandlekar2023mimicgen} expands about 100 teleoperated simulation demonstrations into 10,000 episodes; half include added sinusoidal finger perturbations before grasping. A procedural generator adds 3,700 random-motion episodes where a random subset of finger joints follows random sine waves while the end effector moves between randomly sampled targets. A further 250 episodes are teleoperated directly.

The real dataset comprises 218 episodes (2.5\,h at 5\,fps), teleoperated with a Rokoko glove~\cite{rokoko2025}: 198 short pick-and-place rollouts ($\le$40\,s, 0.71\,h) and 20 long free-play sequences (1.80\,h) in which the operator grasps, releases, pushes, and re-grasps the object without a task goal, almost exclusively with a red cube. Masks are pseudo-labeled with SAM~3. For the third-person view, "hand" and the object name are used as prompts for automatic pseudo-labeling. For the wrist view, a short initialization clip showing the hand moving into frame seeds SAM~3's tracker via its extracted bounding box.

The real demonstrations cover only a narrow range of the action space. Taking the 1st to 99th percentile of each dimension as its covered range, the simulation corpus spans 2.1$\times$ the real range across the 17 hand joints (1.2$\times$ to 7.7$\times$ per joint) and 1.5$\times$ across the end-effector pose, with the largest gap in finger abduction (4.6$\times$ over the abduction joints; the single-joint maximum of 7.7$\times$ is middle-finger abduction). Simulation midtraining exposes the dynamics model to this broader range of hand motions.
\subsection{Baselines and Training Recipes}
\label{sec:evaluation:monolithic}
Table~\ref{tab:model_design} lists five models, named by architecture and training data. \emph{Cascade} denotes our two-stage \modelname{} architecture while \emph{Mono} denotes the monolithic approach. The suffixes \emph{R}, \emph{S}, and \emph{SR} denote training on real data only, simulation data only, and simulation data followed by real data, respectively. For Cascade variants, the suffix refers to \wmone{}; all three share the same \wmtwo{}, trained only on real data.
The monolithic baselines adapt Ctrl-World~\cite{guo2026ctrlworld} to the same 23-D action space using separate MLPs for end-effector pose and hand-joint commands, with their outputs summed before frame-wise cross-attention. \monosr{} uses the same simulation corpus as \msr{}, rendered as RGB; \monoro{} uses only real data.

\begin{table}[t]
\vspace*{4pt}
\caption{Model Design}
\label{tab:model_design}
\centering
\footnotesize
\resizebox{\columnwidth}{!}{%
\begin{tabular}{lcccc}
\toprule
Model & Sim midtr. & Real FT & Mask decomp. & \shortstack{\# gradient updates\\(sim/real)} \\
\midrule
\monoro{} & -- & \checkmark & -- & --\,/\,22k \\
\monosr{} & \checkmark & \checkmark & -- & 55k\,/\,58k \\
\mro{} & -- & \checkmark & \checkmark & --\,/\,15k \\
\mso{} & \checkmark & -- & \checkmark & 55k\,/\,-- \\
\msr{} & \checkmark & \checkmark & \checkmark & 55k\,/\,45k \\
\bottomrule
\end{tabular}}
\par\vspace{2pt}
\parbox{\columnwidth}{\footnotesize \wmtwo{} is shared by all Cascade rows (real data only, 70k steps).}
\end{table}

\label{sec:models:training}
All models use AdamW with a cosine learning-rate schedule; training-step counts are listed in Table~\ref{tab:model_design}. All LoRA adapters use rank 16 and $\alpha{=}16$. \wmone{} is midtrained in simulation (lr $10^{-4}$, batch 64, one H200 GPU), then LoRA-fine-tuned on real data (lr $5{\times}10^{-6}$). \wmtwo{} is trained on real data (lr $10^{-4}$, batch 72, 8$\times$H100, with LoRA). \monosr{} follows \wmone{}'s two-stage recipe with freshly initialized LoRA adapters for real-data fine-tuning. For real-only training from SVD weights, \monoro{} uses LoRA (lr $10^{-4}$, batch 64), while \mro{} updates all \wmone{} weights (lr $10^{-4}$). The real-data stages are not step-matched: each monolithic baseline receives more real-data updates than its corresponding cascade variant. Inference uses 50 denoising steps per autoregressive step.

\subsection{Geometric Controllability}
\label{sec:evaluation:geometry}
\label{sec:results:geometry}
We evaluate controllability in three ways. First, raters compare how closely the generated hand matches the commanded target posture. Second, raters assess whether each commanded degree of freedom moves independently while the others remain fixed. Third, for the mask-prediction models, we measure the overlap between the predicted hand mask and the target mask.

\subsubsection{Protocol}
A trial samples a random 23-D target command and rolls the model out for 10 autoregressive steps (50 frames). The final rendered frame is judged against a simulation-rendered reference of the commanded target posture. To distinguish the test commands from the sinusoidal pattern in the simulation training data, we use step or ramp commands instead. 

\subsubsection{Raters and metrics}
\begin{figure}[t]
  \vspace*{4pt}
  \centering\footnotesize
  \makebox[0.3167\linewidth]{Reference}\makebox[0.3167\linewidth]{Model A}\makebox[0.3167\linewidth]{Model B}\\[1pt]
  \includegraphics[width=0.95\linewidth]{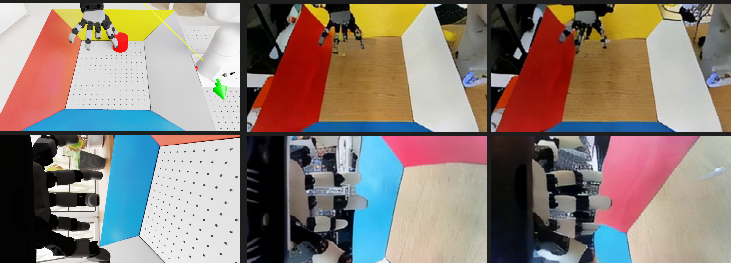}
  \caption{\textbf{Pairwise comparison shown to raters.} Left: the simulation-rendered target reference (raters additionally see it as a hand/arm mask); middle and right: the two blinded candidate rollouts. Third-person view above the wrist view in each panel.}
  \label{fig:ab_test_example}
\end{figure}

Targets are random whole-pose states and single-dimension perturbations. Raters, blind to model identity, see the reference (RGB and hand/arm mask) and the final frames of two candidate rollouts in both views (Fig.~\ref{fig:ab_test_example}), judge the hand configuration only, and answer A, B, or tie; A/B slots are assigned at random. We report the win rate from pairwise comparisons across 50 target scenes.

Second, a sine-sweep study tests single-dimension control directly: each of the 23 dimensions is driven with one sinusoid cycle while the others are held at their initial values, from 10 starting episodes per model (1,150 rollouts). Raters, blind to model identity, see the rollout and a description of the commanded dimension and score it 1 (only that dimension moves), 0.5 (it moves with unintended motion elsewhere), or 0 (no movement). Rollouts are presented in randomized order. A model's score is the mean rating over dimensions, episodes, and raters, with its standard error.

Finally, for the three mask-prediction variants, we evaluate \wmone{}'s predicted hand geometry using the intersection over union (IoU) between its predicted hand mask and the simulator's target mask. We compute IoU on the final frame of separate 25-frame rollouts (5 autoregressive steps), classifying a pixel as hand when its RGB distance to pure green is below 130.

For each model pair, the non-tied votes of their direct comparisons are tested against a fair coin with an exact two-sided binomial test, Bonferroni-corrected~\cite{dunn1961} over the 10 pairs ($\alpha{=}0.005$). We compare IoU using paired Wilcoxon signed-rank tests~\cite{wilcoxon1945}, with all three \wmone{} variants evaluated on the same 560 trials (28 targets $\times$ 2 schedules $\times$ 10 base episodes).

\subsubsection{Results}

\begin{figure}
    \vspace*{4pt}
    \centering
    \includegraphics[width=0.90\linewidth]{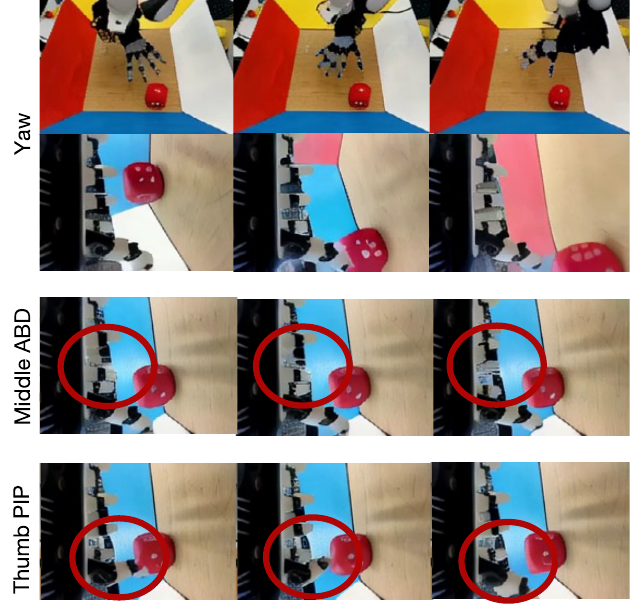}
    \caption{\textbf{Sine-sweep examples (\msr{}).} Three frames each for yaw (third-person view above the wrist view), middle-finger abduction, and thumb PIP flexion; red circles mark the commanded joint. Under yaw the object keeps a consistent appearance across the wrist-camera frames.}
    \label{fig:sine_sweep_examples}
\end{figure}

Table~\ref{tab:controllability} ranks the five models by human pairwise win rate. \msr{} leads by a wide margin. \monoro{}, the strongest monolithic model, sits level with \mso{} and below \mro{}. \monosr{}, the monolithic model trained with \msr{}'s recipe, is last: every \modelname{} variant's 95\% confidence interval~\cite{wilson1927} clears its interval with no overlap.

Head-to-head tests on the non-tied votes give four results. First, decoupling helps at both budgets: every \modelname{} variant beats \monosr{} ($p<10^{-12}$), \msr{} in 94\% of non-tied comparisons, and \mro{} beats \monoro{} in 69\% ($p{=}1.5{\times}10^{-7}$). Second, simulation midtraining has opposite effects: raters prefer \msr{} over \mro{} (73\%, $p{=}7.5{\times}10^{-14}$) and over \mso{} (80\%, $p{=}1.9{\times}10^{-22}$), but \monoro{} over \monosr{} (74\%, $p{=}1.3{\times}10^{-15}$). Third, the full model beats the strongest monolithic model: \msr{} over \monoro{}, 75\% ($p{=}7.8{\times}10^{-17}$). Fourth, simulation without the real-data stage is not enough: \mso{} ties \monoro{} ($p{=}0.32$) and \mro{} ($p{=}0.046$, not significant after correction).

\begin{figure}[t]
  \centering
  \includegraphics[width=0.90\linewidth]{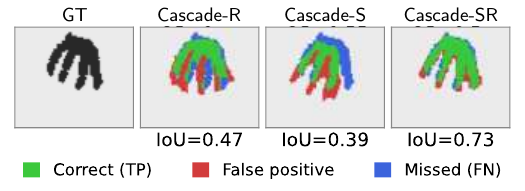}
  \caption{\textbf{Hand-mask IoU example.} Target hand mask (left) and each \wmone{} variant's final-frame prediction overlaid on it (green = correct, red = false positive, blue = missed), with its IoU.}
  \label{fig:wm1_iou_qualitative}
\end{figure}

\begin{table}[t]
\vspace*{4pt}
\caption{Controllability: A/B Win Rate, Sine-Sweep Score, and \wmone{} Hand-Mask IoU}
\label{tab:controllability}
\centering
\footnotesize
\setlength{\tabcolsep}{4pt}
\begin{tabular}{lccc}
\toprule
Model & A/B win rate & Sine-sweep score & \wmone{} IoU \\
\midrule
\monoro{} & 35.3\% [32.8, 37.8] & 0.77 $\pm$ 0.02 & -- \\
\monosr{} & 16.4\% [14.6, 18.4] & 0.75 $\pm$ 0.02 & -- \\
\mro{} & 40.8\% [38.3, 43.4] & 0.79 $\pm$ 0.02 & 0.412 \\
\mso{} & 35.3\% [32.9, 37.8] & 0.81 $\pm$ 0.02 & 0.414 \\
\msr{} & \textbf{63.0\% [60.5, 65.5]} & \textbf{0.87 $\pm$ 0.01} & \textbf{0.447} \\
\bottomrule
\end{tabular}
\par\vspace{2pt}
\parbox{\columnwidth}{\footnotesize A/B win rate [95\% Wilson interval~\cite{wilson1927}] over all comparisons with the other four models, ties as non-wins (3,549 ratings). Sine-sweep score: mean $\pm$ standard error of the 0/0.5/1 ratings over dimensions, episodes, and raters. IoU: \wmone{}'s predicted hand mask against the simulation-rendered target on the final frame of separate 25-frame rollouts.}
\end{table}

The sine-sweep ratings (Table~\ref{tab:controllability}, third column; Fig.~\ref{fig:sine_sweep_examples}) give the same top of the ranking: \msr{} scores 0.87, about three standard errors of the difference above \mso{} (0.81), while \mso{}, \mro{}, \monoro{}, and \monosr{} (0.81 to 0.75) lie within about two standard errors of each other.

For hand-mask IoU (Table~\ref{tab:controllability}; Fig.~\ref{fig:wm1_iou_qualitative}), \msr{} exceeds \mro{} by $\Delta$IoU $= {+}0.035$ (bootstrap~\cite{efron1993bootstrap} 95\% CI $[0.029, 0.042]$, $p{=}2{\times}10^{-22}$) and \mso{} by ${+}0.034$ ($[0.023, 0.044]$, $p{=}3{\times}10^{-8}$), while \mro{} and \mso{} tie ($p{=}0.44$).

\subsection{Video and Perceptual Quality}
\label{sec:results:video}
\label{sec:evaluation:video}
\subsubsection{Protocol}
We assess video quality on held-out evaluation episodes, reporting results separately for each camera view (Table~\ref{tab:results_video}). For frame-level fidelity, we report peak signal-to-noise ratio (PSNR), which measures pixel reconstruction fidelity in decibels; the structural similarity index (SSIM), which compares local luminance, contrast, and structure; and learned perceptual image patch similarity (LPIPS)~\cite{zhang2018lpips}, which measures perceptual differences using deep features. To assess distributional similarity between predictions and ground truth, we report Fr\'echet inception distance (FID)~\cite{heusel2017fid} on image features and Fr\'echet video distance (FVD)~\cite{unterthiner2018fvd} on spatiotemporal video features. Higher PSNR and SSIM and lower LPIPS, FID, and FVD indicate better quality. We also measure sharpness using the Laplacian variance~\cite{pechpacheco2000diatom} of predicted frames relative to ground truth (1 = as sharp as ground truth).

\subsubsection{Results}
\msr{} is on par with the strongest monolithic model (Table~\ref{tab:results_video}). In the third-person view it is within 0.3 PSNR of \monoro{} with higher SSIM; in the wrist view it trails by 0.9 PSNR and 0.014 LPIPS; it is better on FVD and FID in both views. Against \monosr{} it is better on every metric in both views ($+1.9$ and $+0.6$ PSNR). 
\mro{} achieves the best video-quality metrics in both views but has lower controllability scores than \msr{} (see Sec.~\ref{sec:evaluation:geometry}); this trade-off may partly reflect its full-weight training of \wmone{} on real data alone.

Simulation midtraining also reduces perceptual quality in the monolithic model: \monosr{} scores worse than \monoro{} on every pixel and distribution metric in both views, despite receiving more real-data training steps (Table~\ref{tab:model_design}). Its third-person sharpness is 0.90 times that of ground truth, compared with more than 0.95 for every other model.

Wrist-view PSNR is 5.9--7.2\,dB lower than third-person PSNR for every model. The wrist camera moves with the hand. Under end-effector motion with fixed finger joints, the hand silhouette remains approximately fixed while the surrounding scene moves. The three-class mask does not encode motion within the background region; action conditioning supplies \wmtwo{} with additional motion information. Consistent with this, removing \wmtwo{}'s action conditioning costs 0.024 LPIPS in the wrist view but 0.003 in the third-person view. The sim-midtrained variants also have lower wrist-view mask PSNR in \wmone{} (16.8 for \msr{}, 14.1 for \mso{}, 18.5 for \mro{}). This deficit may reflect differences in camera motion between simulation and reality; it remains present in \msr{} after LoRA fine-tuning.

The monolithic models' pixel scores coincide with blur: in the wrist view \monoro{} and \monosr{} render at 0.87$\times$ and 0.88$\times$ ground-truth sharpness against 0.96$\times$ for \msr{} and 0.99$\times$ for \mro{}, and \msr{} is sharper than \monosr{} in 93\% of samples and than \monoro{} in 87\% of wrist-view samples ($p<10^{-10}$). \mso{} is as blurry as \monoro{} in the wrist view, while \msr{} produces sharper frames after real-data fine-tuning. Pixel metrics can favor averaged predictions~\cite{mathieu2016deep}.

\begin{table*}[t]
\vspace*{4pt}
\caption{Video Quality and Sharpness per Camera View}
\label{tab:results_video}
\centering
\footnotesize
\setlength{\tabcolsep}{4pt}
\resizebox{\textwidth}{!}{%
\begin{tabular}{l cccccc cccccc}
\toprule
& \multicolumn{6}{c}{Third-person view} & \multicolumn{6}{c}{Wrist view} \\
\cmidrule(lr){2-7}\cmidrule(lr){8-13}
Model & PSNR$\uparrow$ & SSIM$\uparrow$ & LPIPS$\downarrow$ & FVD$\downarrow$ & FID$\downarrow$ & Sharp.$\uparrow$ & PSNR$\uparrow$ & SSIM$\uparrow$ & LPIPS$\downarrow$ & FVD$\downarrow$ & FID$\downarrow$ & Sharp.$\uparrow$ \\
\midrule
\monoro{} & \underline{24.08}\,$\pm$\,1.56 & 0.834\,$\pm$\,0.028 & \underline{0.052}\,$\pm$\,0.018 & 6.24 & 14.85 & 0.955 & \underline{17.42}\,$\pm$\,1.07 & \underline{0.559}\,$\pm$\,0.040 & \underline{0.226}\,$\pm$\,0.045 & 11.71 & 26.52 & 0.872 \\
\monosr{} & 21.80\,$\pm$\,1.95 & 0.770\,$\pm$\,0.041 & 0.084\,$\pm$\,0.027 & 8.82 & 24.20 & 0.897 & 15.94\,$\pm$\,1.01 & 0.483\,$\pm$\,0.041 & 0.285\,$\pm$\,0.050 & 13.99 & 31.77 & 0.880 \\
\mro{} & \textbf{24.11}\,$\pm$\,1.89 & \textbf{0.857}\,$\pm$\,0.031 & \textbf{0.045}\,$\pm$\,0.022 & \textbf{4.65} & \textbf{11.81} & \underline{0.978} & \textbf{17.64}\,$\pm$\,1.41 & \textbf{0.587}\,$\pm$\,0.049 & \textbf{0.214}\,$\pm$\,0.057 & \textbf{9.13} & \textbf{22.03} & \textbf{0.994} \\
\mso{} & 21.74\,$\pm$\,1.98 & 0.779\,$\pm$\,0.047 & 0.079\,$\pm$\,0.028 & 7.67 & 16.78 & 0.966 & 14.78\,$\pm$\,0.96 & 0.443\,$\pm$\,0.046 & 0.318\,$\pm$\,0.052 & 13.65 & 41.68 & 0.884 \\
\msr{} & 23.74\,$\pm$\,1.86 & \underline{0.843}\,$\pm$\,0.033 & 0.059\,$\pm$\,0.026 & \underline{5.30} & \underline{13.39} & \textbf{0.979} & 16.54\,$\pm$\,1.23 & 0.548\,$\pm$\,0.045 & 0.240\,$\pm$\,0.054 & \underline{9.60} & \underline{23.32} & \underline{0.956} \\
\bottomrule
\end{tabular}}
\par\vspace{2pt}
\parbox{\textwidth}{\footnotesize 150 held-out samples, 50-frame rollouts. PSNR, SSIM, LPIPS: mean $\pm$ standard deviation over samples. FVD (I3D, 16 frames per clip, 150 clips per view) and FID (Inception-v3, 8 frames per clip, 1,200 frames per view) are dataset-level Fr\'echet distances computed per view; they carry no per-sample spread. Sharp.: Laplacian variance of the predicted frames relative to ground truth (1 = as sharp as ground truth)~\cite{pechpacheco2000diatom}. Bold = best, underline = second best per column.}
\end{table*}

\subsection{Qualitative Failure Modes}
\label{sec:results:failures}
\begin{figure*}[t]
  \centering
  \includegraphics[width=0.8\linewidth]{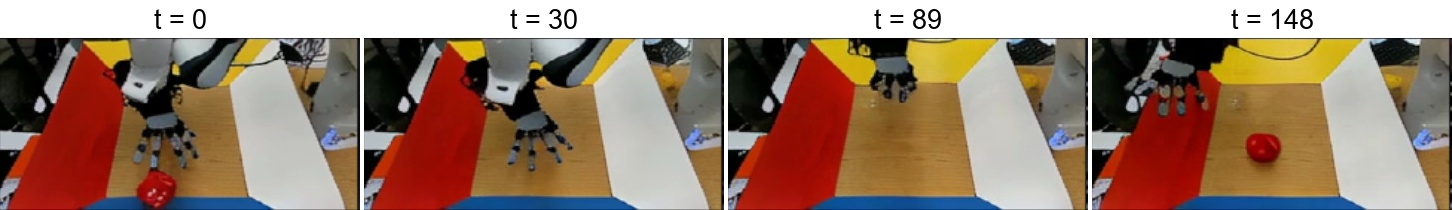}\\[0.3em]
  \includegraphics[width=0.8\linewidth]{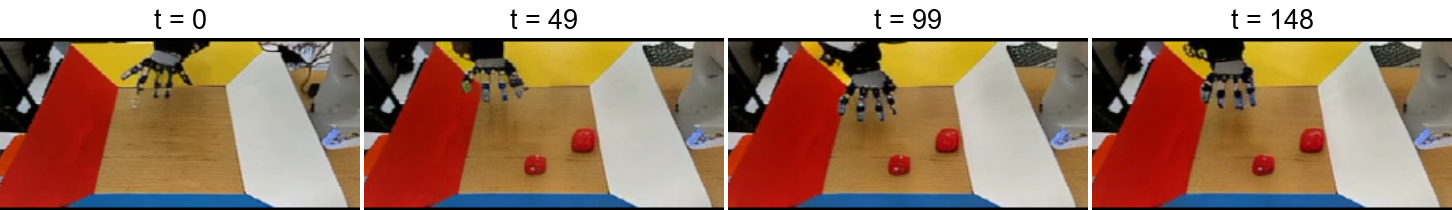}
  \caption{\textbf{Object-permanence failure modes.} Third-person generated frames from 149-frame rollouts (5 frames predicted per step). Top: object vanishing under occlusion -- present early, disappears while occluded by the hand, and reappears once the occlusion clears. Bottom: object duplication/spawning -- \wmone{} predicts an incorrect object state, causing a second cube to appear or the object to shift location.}
  \label{fig:failure_modes}
\end{figure*}

Two recurring failure modes originate in \wmone{}'s object-state prediction (Fig.~\ref{fig:failure_modes}): under hand occlusion the object mask is sometimes dropped, leaving an empty scene, and the object is occasionally duplicated or respawned elsewhere. Both show that object permanence is what limits long autoregressive deployment.

\section{Discussion and Limitations}
\label{sec:discussion}

\subsection{Interpreting the Comparison}
The comparisons in Section~\ref{sec:results:geometry} show improved controllability with mask decomposition under both real-only training and simulation midtraining followed by real-data fine-tuning. Simulation midtraining further benefits the mask dynamics model but reduces controllability in the monolithic model. The simulation uses no material, lighting, or texture tuning (Section~\ref{sec:evaluation:setup}); the appearance gap between simulated and real RGB may contribute to the monolithic model's degradation. These results support learning dynamics in mask space, with real RGB supervision reserved for the rendering model. The full model combines stronger controllability with comparable perceptual quality and sharper frames than the monolithic baselines.

\subsection{Limitations and Future Work}

\label{sec:limitations}
\emph{Evaluation scope.} This study uses a single physical scene with data collected on a single pick-and-place task. Robustness to distribution shift, such as new objects or scene changes, is left to future work. Moreover, the effect of the improved controllability on downstream tasks like planning or policy evaluation has not been tested. 

\emph{Judging.} The pooled tests treat every vote as an independent trial and do not model variation between raters. The nonsignificant differences between \mro{} and \mso{}, and between \mso{} and \monoro{}, should be interpreted cautiously; they do not establish equivalence.

\emph{Pseudo-Labeling.} The segmentation-mask supervision relies on SAM~3 pseudo-labels, which may contain boundary or classification errors. This effect is most pronounced for \wmtwo{}, which is trained on real data only.

\emph{Model.} \wmone{} receives no depth input and drops occluded objects (Section~\ref{sec:results:failures}); depth conditioning is a natural remedy. A background or depth channel added to the mask vocabulary would also give \wmtwo{} the egomotion cue it currently lacks in the wrist view (Section~\ref{sec:results:video}). The color-class mask does not encode object identity. Object-ID or text-prompt conditioning could help preserve identity over long horizons. A natural next baseline pairs a kinematically rendered (forward-kinematics) hand mask with a learned object-mask predictor.

\emph{Future work.} Other directions include conditioning on camera intrinsics and extrinsics to remove the fixed-viewpoint assumption. Moreover, two sequential diffusion passes are computationally expensive; distillation into few-step samplers is the natural route to real-time rollouts.
\section{Conclusion}
\label{sec:conclusion}
We introduced \modelname{}, a two-stage action-conditioned world model for 23-DoF dexterous manipulation that decouples prediction into mask dynamics and RGB rendering. Training the dynamics model on a larger and more diverse simulation dataset further improves controllability. Raters prefer the full model over monolithic baselines in blind A/B comparisons with random target commands and assign it higher action-following scores in the single-DoF actuation experiment. \modelname{} performs on par with the strongest monolithic model on  perceptual metrics, and yields generally sharper predictions. These results suggest that decoupling dynamics from rendering is a promising path toward action-conditioned world models that are both controllable and data-efficient for highly dexterous embodiments.

\section*{Acknowledgment}
\enlargethispage{-0.3cm}
Generative AI tools (large language models) assisted in preparing this work: drafting and revising text under the authors' direction, and writing and debugging evaluation code. All ideas, methods, and interpretations are the authors' own, who reviewed all AI-assisted content and take full responsibility for it.

\bibliographystyle{IEEEtran}
\bibliography{references}

\end{document}